%% file: main.tex
\documentclass[10pt,twocolumn,letterpaper]{article}

\usepackage{cvpr}              

\usepackage{caption}
\usepackage{multirow}
\usepackage{booktabs}
\usepackage{slashed}
\definecolor{cvprblue}{rgb}{0.21,0.49,0.74}
\usepackage[pagebackref,breaklinks,colorlinks,allcolors=cvprblue]{hyperref}

\def\paperID{} 
\def\confName{CVPR}
\def\confYear{2026}

\usepackage{xcolor}

\title{SharpTimeGS: Sharp and Stable Dynamic Gaussian Splatting \\via Lifespan Modulation}

\author{Zhanfeng Liao$^{1}$, Jiajun Zhang$^{2}$, Hanzhang Tu$^{1}$, Zhixi Wang$^{1}$, Yunqi Gao$^{4}$, \\ Hongwen Zhang$^{3}$, Yebin Liu$^{*, 1}$\\
$^1$Tsinghua University $^2$Beijing University of Posts and Telecommunications \\ 
$^3$Beijing Normal University $^4$Central China Normal University
}

\begin{document}
 \maketitle
 \let\thefootnote\relax\footnotetext{$^*$Corresponding author.}
 \input{sec_0_abstract}
 \input{sec_1_intro}
 \input{sec_2_relatework}
 \input{sec_3_method}
 \input{sec_4_experiments}
 \input{sec_5_conclusion}
\newpage
{
    \small
    \bibliographystyle{ieeenat_fullname}
    \bibliography{main}
}


\end{document}

%% file: sec_0_abstract.tex
\begin{abstract}
Novel view synthesis of dynamic scenes is fundamental to achieving photorealistic 4D reconstruction and immersive visual experiences. 
Recent progress in Gaussian-based representations has significantly improved real-time rendering quality, yet existing methods still struggle to maintain a balance between long-term static and short-term dynamic regions in both representation and optimization. 
To address this, we present SharpTimeGS, a lifespan-aware 4D Gaussian framework that achieves temporally adaptive modeling of both static and dynamic regions under a unified representation.
Specifically, we introduce a learnable lifespan parameter that reformulates temporal visibility from a Gaussian-shaped decay into a flat-top profile, allowing primitives to remain consistently active over their intended duration and avoiding redundant densification. 
In addition, the learned lifespan modulates each primitive's motion, reducing drift in long-lived static points while retaining unrestricted motion for short-lived dynamic ones. This effectively decouples motion magnitude from temporal duration, improving long-term stability without compromising dynamic fidelity.
Moreover, we design a lifespan-velocity-aware densification strategy that mitigates optimization imbalance between static and dynamic regions by allocating more capacity to regions with pronounced motion while keeping static areas compact and stable.
Extensive experiments on multiple benchmarks demonstrate that our method achieves state-of-the-art performance while supporting real-time rendering up to 4K resolution at 100 FPS on one RTX 4090. Project page: https://liaozhanfeng.github.io/SharpTimeGS.

\end{abstract}

%% file: sec_1_intro.tex
\vspace{-6mm}
\section{Introduction}
\label{sec:intro}
Dynamic novel view synthesis (NVS) aims to generate photorealistic free-viewpoint videos of dynamic scenes, enabling immersive applications in VR/AR, telepresence, and digital content creation.
Traditional geometry-based pipelines, such as multi-view stereo~\cite{schonberger2016structure,furukawa2009accurate}, mesh-based capture~\cite{casas20144d,collet2015high}, and volumetric fusion~\cite{dou2016fusion4d}, explicitly recover scene geometry but require dense camera rigs and controlled environments.
Image-based view interpolation~\cite{lipski2009virtual} sidesteps explicit reconstruction, yet struggles to maintain geometric and temporal consistency.
Neural Radiance Fields~\cite{mildenhall2020nerf,park2021nerfies,park2021hypernerf,pumarola2021d,li2022neural,fridovich2023k,cao2023hexplane} alleviate these limitations through continuous implicit representations that achieve high rendering fidelity. However, their heavy computation and slow rendering speed limit practical deployment for dynamic scenes.

Recently, 3D Gaussian Splatting (3DGS)~\cite{kerbl3Dgaussians} has emerged as a highly efficient explicit representation, enabling real-time photorealistic rendering. Extending 3DGS to dynamic scenes has led to two major paradigms. 
Canonical-space deformation methods~\cite{wu20244d,yang2024deformable, liang2025gaufre} learn per-frame deformation fields that warp a static canonical representation to each timestamp. 
Although conceptually appealing, these approaches struggle with complex or large-scale motions because of the difficulty of optimizing high-dimensional deformation fields and maintaining temporal coherence. On the other hand, motion-based methods like 4DGS~\cite{yang2024real}, 4DRotorGS~\cite{duan20244d}, STGS~\cite{li2024spacetime}, and FreeTimeGS~\cite{wang2025freetimegs} model the time-varying motion of Gaussian primitives in 3D space.

\begin{figure}[t!]
    \centering
    \includegraphics[width=1.0\linewidth]{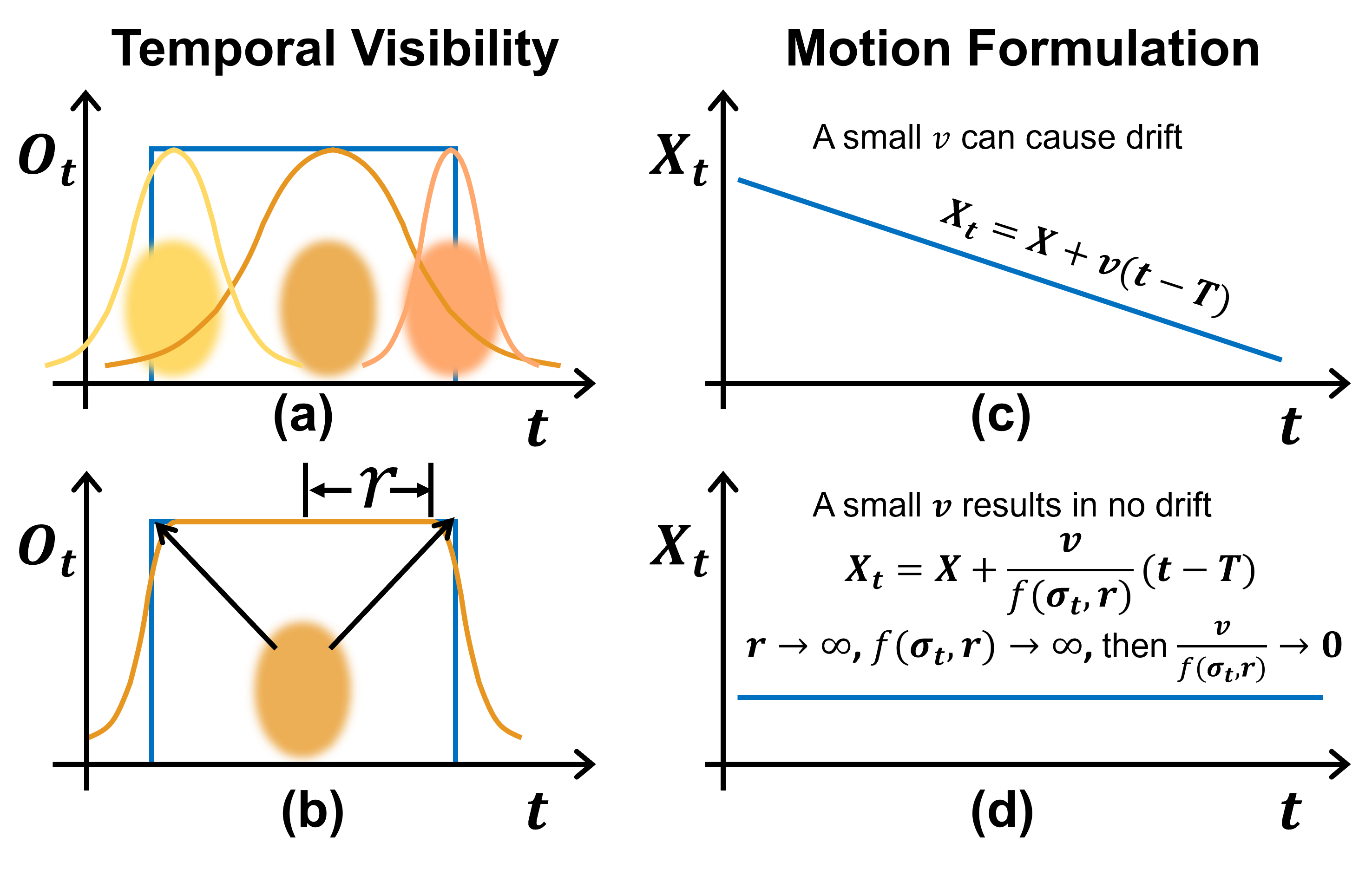}
    \vspace{-9mm}
    \caption{
    (a) Temporal visibility in existing motion-based methods. A step-like lifespan (blue line) requires multiple Gaussian primitives for approximation.
    (b) With a learnable radius $r$, our visibility function allows a single Gaussian primitive to represent a step-like lifespan (blue line).
    (c) In existing motion-based methods (e.g., FreeTimeGS~\cite{wang2025freetimegs}), residual velocities accumulate over time, causing drift in static regions.
    (d) With our lifepan modulation term $f(\sigma_t, r)$, where $\sigma_t$ is the lifespan variance and $r$ is the lifespan radius, static primitive remain static without drift.
    }
    \vspace{-9mm}
    \label{fig:teaser}
\end{figure}

However, the temporal visibility profile and motion formulation for Gaussian primitives used in these methods overlook the fundamental differences between static and dynamic points.
Specifically, for temporal visibility, existing methods use a Gaussian curve to model opacity (Fig.\ref{fig:teaser}(a)). However, its bell-shaped profile causes long-lived primitives to decay gradually. As a result, representing a flat, time-invariant visibility requires multiple overlapping Gaussians to approximate the true curve, leading optimization to repeatedly insert new primitives (Fig.\ref{fig:teaser}(a), other primitives).
For motion modeling, existing methods neglect to fully model the relationship between velocity and lifespan.
As a consequence, a static primitive must learn an extremely small velocity to remain stable. 
However, it is practically impossible for optimization to converge to an absolute zero velocity, so even tiny residual motion inevitably accumulates over long time periods and leads to noticeable spatial drift and instability (Fig.\ref{fig:teaser}(c)).
In essence, while a behavior-agnostic formulation brings simplicity and consistent optimization, it inevitably entangles static and dynamic behaviors, making it difficult to represent both faithfully within a single representation. 

To address these challenges, we propose SharpTimeGS, a novel 4D Gaussian representation that makes temporal behavior adaptive through a learnable lifespan, enabling balanced modeling of long-term static structures and short-lived dynamic motions without breaking the unified representation.
Our key insight is that the characteristics of a primitive’s motion and temporal visibility are strongly tied to its lifespan.
We therefore introduce lifespan as a learnable per-primitive attribute and incorporate it directly into the opacity and motion formulations.
First, we reformulate temporal visibility by replacing the conventional Gaussian decay with a flat-top profile that controls the lifespan. This design eliminates the smooth, gradual decay characteristic of Gaussian kernels, allowing primitives to maintain stable opacity within their active lifespan and to drop off sharply once that lifespan ends (Fig.~\ref{fig:teaser}(b)), thereby removing motion dragging and producing clearer temporal boundaries. 
Second, by using lifespan as a modulation term in motion formulation, long-lived primitives naturally suppress displacement and remain stable (Fig.~\ref{fig:teaser}(d)) because the lifespan-induced scaling attenuates their effective motion, whereas short-lived primitives preserve expressive motions. This enables a unified motion formulation that balances static stability and dynamic flexibility.

Based on the lifespan modulation representation, we further perform dynamic–static separation at initialization, assigning short lifespans and initial velocities to dynamic primitives while keeping static ones long-lived and stationary, which significantly stabilizes optimization.
Finally, to balance optimization between static and dynamic regions, we introduce a lifespan–velocity–aware densification strategy. Each primitive is assigned a score reflecting its motion magnitude relative to its lifespan, allowing short-lived, fast-moving primitives to be cloned more frequently while long-lived static ones remain compact. This adaptive allocation assigns more representational capacity to dynamic regions while keeping static areas stable and lightweight. 
Together, these components yield a unified 4D Gaussian representation that remains stable over time and faithfully captures dynamic motion.

Extensive experiments on Neural3DV\cite{li2022neural}, ENeRF-Outdoor\cite{lin2022efficient}, and SelfCap~\cite{xu2024representing} datasets demonstrate that SharpTimeGS achieves state-of-the-art dynamic scene rendering quality compared with existing methods. The contributions of this work are summarized as follows:
\begin{itemize}
    \item A lifespan-controlled flat-top visibility that avoids the gradual Gaussian decay, thereby removing motion dragging and producing sharper temporal boundaries.
    \item A lifespan-modulated motion formulation that balances static stability and dynamic expressiveness.
    \item A velocity-aware initialization that separates dynamic and static points and assigns lifespan–velocity priors accordingly, stabilizing the optimization process.
    \item A lifespan–velocity–aware densification strategy that prioritizes dynamic regions while keeping static areas compact and stable.
\end{itemize}

%% file: sec_2_relatework.tex
\vspace{-4mm}
\section{Related Work}
\label{sec:relatedwork}
\vspace{-2mm}
\paragraph{NeRF-based dynamic NVS.}
With the advent of NeRF~\cite{mildenhall2020nerf} and differentiable rendering, neural scene representations have become the mainstream paradigm for dynamic scene reconstruction. Extending static NeRF-related representations to dynamic scenes are also being actively explored ~\cite{attal2023hyperreel,park2021nerfies,park2021hypernerf,pumarola2021d,li2022neural,fridovich2023k,cao2023hexplane,shao2023tensor4d,icsik2023humanrf,kim2024sync,peng2023representing,NeRFPlayer,wang2023mixed,wang2023neural}.
Methods such as NeRFies~\cite{park2021nerfies}, HyperNeRF~\cite{park2021hypernerf}, and D-NeRF~\cite{pumarola2021d} construct a canonical NeRF and learn per-frame deformation fields via multi-layer perceptrons (MLPs).
NeRFies~\cite{park2021nerfies} linked the observation space to a canonical space via deformation fields.
Neural3DV~\cite{li2022neural} employs time-conditioned neural fields to directly represent dynamics in 4D space, offering strong expressiveness but incurring heavy computational and memory costs.
To improve efficiency and scalability, hybrid representations like K-Planes~\cite{fridovich2023k}, HEX-Plane~\cite{cao2023hexplane}, and Tensor4D~\cite{shao2023tensor4d} combine voxel grids with neural fields, significantly reducing training and inference time.
HEX-Plane~\cite{cao2023hexplane} decomposes the 4D domain into six feature planes, where point features are sampled via interpolation and concatenated to predict density and color. Similarly, K-planes~\cite{fridovich2023k} offers a unified approach for both static and dynamic scenes.
Despite these advances, NeRF-based methods still face challenges in dynamic scene reconstruction, including slow rendering, limited quality, and high storage overhead.
In contrast, our approach achieves superior rendering efficiency, quality, and scalability, making it a practical solution for large-scale dynamic scene reconstruction.

\vspace{-3mm}
\paragraph{Gaussian-based dynamic NVS.}
3D Gaussian Splatting (3DGS)~\cite{kerbl3Dgaussians} has emerged as a competitive alternative for dynamic scenes, offering real-time rendering with sharp detail. 
Approaches to modeling motion with Gaussian Splatting can be broadly
categorized into two types: deformation-based and motion-based methods.
Deformation-based methods~\cite{yang2024deformable,bae2024per,qingming2025modgs,guo2024motion,lu20243d,shaw2024swings,zhu2024motiongs,labe2024dgd,liang2025gaufre,xu2024grid4d,kim20244d,4k4d} employ MLPs or low-rank K-planes to dynamically adjust the parameters of Gaussians over time. 
While expressive, the continuous-deformation formulation introduces overhead that hampers training speed and rendering efficiency.
Deformable-3DGS~\cite{yang2024deformable} applies forward deformation, warping each Gaussian primitive from canonical space to observation space before rendering.
More recent methods~\cite{yang2024real,lee2024fully,luiten2024dynamic,xu2024representing,wang2025freetimegs,li2024spacetime,duan20244d,gao20257dgs} eschew deformations and directly model dynamics with 4D Gaussian primitives.
%
In 4DGS~\cite{yang2024real} and 4DRotorGS~\cite{duan20244d}, geometry and velocity are entangled, complicating joint optimization.
The spatial and temporal scales are tightly coupled, yielding complex convergence behavior.
STGS~\cite{li2024spacetime} employs polynomial motion with angular velocity, but the high-order, high-dimensional parameterization is difficult to optimize in complex regimes and prone to overfitting. 
FreeTimeGS~\cite{wang2025freetimegs} adopts linear velocities but lacks a unified treatment of static versus dynamic regions, leading to static jitter and loss of detail in fast motions.
7DGS~\cite{gao20257dgs} introduces a unified representation over position, time, and view direction by slicing 7D Gaussians into 3D subspaces, similar to 4DGS. Nevertheless, it still struggles to model motion-related dynamics.

%% file: sec_3_method.tex
\vspace{-3mm}
\section{Method}
\vspace{-2mm}
\label{method}
\begin{figure*}[ht]
    \centering
    \includegraphics[width=1.0\linewidth]{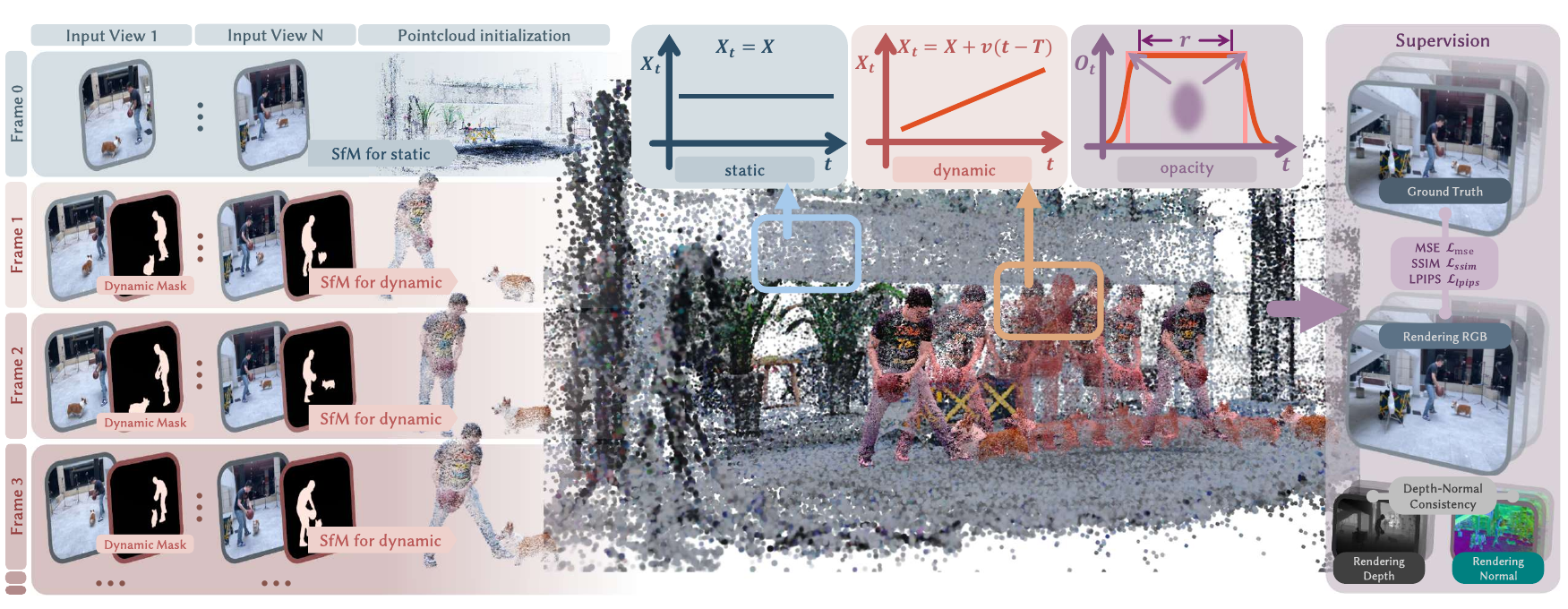}
    \vspace{-7mm}
    \caption{The pipeline of our method. We represent a dynamic scene using Gaussian primitives whose temporal visibility adapts to the actual lifespan of each point. To achieve this, we introduce a lifespan-dependent parameter $r$ that modulates the temporal Gaussian, allowing a single primitive to accurately model its full lifespan. Moreover, through the modulation terms $f(\sigma_t, r)$ related to $\sigma_t$ and $r$, the static part can be completely static and still able to express dynamic parts (the static and fast dynamic regions will be transformed into equations of motion in red and blue boxes, respectively). Note that the formulas in the boxes are only approximations. During optimization, all Gaussian representations remain identical.}
    \label{fig:framework}
    \vspace{-6mm}
\end{figure*}
Given multi-view videos of a dynamic scene, our goal is to reconstruct a temporally continuous 4D representation for novel view synthesis over time.
Fig.~\ref{fig:framework} demonstrates the overall pipeline of SharpTimeGS.
The framework consists of a velocity-aware initialization that provides motion priors (Sec.~\ref{initialization}), a lifespan-modulated 4D Gaussian representation that adjust temporal visibility and motion (Sec.~\ref{modeling}), and a velocity–lifespan–aware densification strategy (Sec.~\ref{densification}). 
\subsection{4D Gaussians with Lifespan Modulation}
\label{modeling}
The vanilla 3DGS~\cite{kerbl3Dgaussians} formulation models static scenes using a set of spatial Gaussian primitives, each parameterized by its position $X$, scale $S$, rotation $R$, opacity $O$ and view-dependent color coefficients $Y$. To extend this representation to dynamics scenes, recent works~\cite{yang2024real,li2024spacetime,wang2025freetimegs,lee2024fully} introduce a temporal dimension by assigning each primitive a time anchor $T$ and modeling its motion and opacity as functions of the time offset $\Delta t = t- T$. 
These formulations currently omit explicit lifespan modeling, even though both velocity and temporal opacity are inherently lifespan-dependent. 
To address this limitation, we incorporate a lifespan parameter to enhance both the efficiency of expression and optimization.
%

\vspace{-4mm}
\paragraph{Lifespan-Modulated Motion Dynamics.}
We observe that a primitive’s motion behavior is inherently tied to its temporal lifespan.
Based on this, we formulate a lifespan-modulated motion function that adaptively scales motion amplitude by each primitive's temporal persistence, defined as:
\begin{equation}
\label{eqn:deform}
\begin{aligned}
 {X_t} &= {X} + \frac{v}{f(\sigma_t, r)}(t-T),\\
 f(\sigma_t, r) &= 1.0 + \text{max}\left\{1.0, (\sigma_t+r)^2\right\},
\end{aligned}
\end{equation}
Here, $f(\sigma_t, r)$ denotes the coupled lifespan that modulates the effective motion strength. Specifically, $\sigma_t$ represents the lifespan variance, controlling how gradually a primitive fades over time, and $r$ defines its temporal radius, within which it remains fully active.  
For static regions, $\sigma_t + r$ is large, making $f(\sigma_t, r) \rightarrow \infty$ and thus $v / f(\sigma_t, r) \rightarrow 0$, 
which effectively freezes the primitive’s position over time. 
For dynamic regions with short lifespans, $f(\sigma_t, r)$ becomes small, allowing larger motion amplitudes 
that quickly adapt to rapid changes.
Both $\sigma_t$ and $r$ are learnable parameters for each Gaussian primitive, allowing the model to adaptively adjust temporal behavior during optimization. This proposed formulation naturally balances static stability and dynamic flexibility, enabling both to be represented within the same unified 4D Gaussian space.

\vspace{-4mm}
\paragraph{Lifespan-Modulated Temporal Visibility.}
We further differentiate static and dynamic behaviors at the rendering level by modulating opacity with the same lifespan parameters $r$ used in motion modeling.
The key observation is that most existing approaches still describe the lifespan function using a Gaussian-shaped temporal profile. While this bell-shaped formulation effectively models short-lived dynamic events, it is suboptimal for long-term regions. 
Ideally, static primitives should maintain a flat, time-invariant visibility rather than a peaked Gaussian decay. 
To this end, we reformulate the lifespan function into a flat-top profile that better reflects long-term visibility, defined as:
\begin{gather}
\label{eqn:opacity}
 O_t = O \cdot l(t), \notag \\
 l(t) =
\begin{cases}
\exp{\left(-\left(\dfrac{\left|t - T\right| - r}{\sigma_t}\right)^2\right)}, & \left|t - T\right| > r, \\
1, & \left|t - T\right| \le r,
\end{cases}
\end{gather}
where $O_t$ denotes the time-dependent opacity, and 
$l(t)$ represents the lifespan modulation function. This design allows static primitives to maintain a stable, time-invariant opacity (large $r$), while dynamic primitives fade in and out over shorter temporal spans (small $r$ and $\sigma_t$), thus unifying both behaviors within a single continuous formulation. 

Based on the moved Gaussian primitive, we calculate its color at position $X_t$ through the spherical harmonics model:
\begin{equation}
C_t = \sum_{l=0}^{L}\;\sum_{m=-l}^{l} C_{lm}\, Y_{lm}\!\left(\text{d}(X_t)\right),
\end{equation}
where $C_t$ is the color of Gaussian primitive at time $t$. $L$, $C_{lm}$, $\text{d}(X_t)$, and $Y_{lm}(\cdot)$ are the degree of spherical harmonics, the spherical harmonics coefficients, the view direction from the center of the camera to the position $X_t$, and the spherical harmonics basis function, respectively.

Finally, we convert the 4D representation to 3D Gaussian primitives at time $t$ and render them following 3DGS~\cite{kerbl3Dgaussians}.

\label{optimization}
\subsection{Velocity-lifespan-aware densification.}
\label{densification}
Fast, complex regions are represented by short-lived, high-velocity Gaussians that receive far fewer effective updates during training than long-lived primitives, leading to blurred or under-detailed reconstructions in dynamic regions.
To alleviate this, we introduce a velocity–lifespan–aware densification strategy that adaptively adjusts Gaussian densification based on motion speed and temporal persistence, achieving a balanced optimization between static and dynamic regions.

We adopt a two-stage training scheme to progressively refine the 4D representation. In the initial stage (the first $1/3$ of training iterations), we follow the densification procedure of AbsGS~\cite{ye2024absgs}, which selects Gaussian primitives based on both average and absolute average image gradients and clones them following \cite{rota2024revising}. The first stage expands the number of Gaussians to sufficiently cover the scene content. When this stage ends, the number of primitives at that point is recorded as $N$, and kept fixed thereafter, while the subsequent stage focuses on refining their spatial-temporal distribution.  

In the second stage, we remove primitives with low opacity and clone new ones based on their velocity–lifespan behavior. To this end, we introduce a scoring metric $s$ for each Gaussian primitive, defined as:
\begin{equation}
\begin{aligned}
 s = \lambda_e E + \lambda_oO + \lambda_l\left(1 - \exp\left(-\frac{\Vert v\Vert+1}{f(\sigma_t, r)}\right)\right).
 \end{aligned}
\end{equation}
This score integrates multiple factors: $E$ denotes the accumulated reconstruction error from rendered and ground-truth images, reflecting how well a primitive explains observed data. The detailed computation of $E$ is provided in the supplementary material.
$O$ represents the opacity of the Gaussian primitive, encouraging preservation of visually significant regions.
The last term prioritizes short-lived, fast-moving primitives by assigning higher scores to Gaussians with large motion magnitude and short lifespan, effectively emphasizing fast and transient motions.
$\lambda_e$, $\lambda_o$, and $\lambda_l$ are weights for these three components.

At each densification step, primitives with opacity below a small threshold are removed.
The same number of new Gaussians are cloned from the top-ranked primitives according to their scores $s$.
This replacement strategy adaptively allocates more representational capacity to transient motions to reconstruct details.

\subsection{Velocity-aware initialization.}
\label{initialization}
A well-designed initialization is crucial for stabilizing 4D Gaussian optimization, especially in dynamic scenes. To provide physically meaningful priors for both spatial and temporal attributes, we design a velocity-aware initialization that separately handles dynamic and static regions.

For the dynamic region, we first identify moving objects by computing optical flow using RAFT~\cite{wang2024sea}. The detected motion points are served as prompts for SAM2~\cite{ravi2024sam} to obtain complete object masks. Given these masks and camera parameters, we reconstruct per-frame 3D point clouds of moving objects using COLMAP~\cite{schonberger2016structure}.
Then, the corresponding points across adjacent frames are matched via K-nearest neighbors (KNN), and their 3D displacements define the initial velocity 
$v_\text{init}$ of Gaussian primitives.
Each extracted dynamic point cloud then initializes standard 3D Gaussian attributes, including position $X$, scale $S$, rotation $R$, opacity $O$, and SH coefficients $Y$, following 3DGS~\cite{kerbl3Dgaussians}. 
The temporal parameters are assigned as follows: the time $T$ corresponds to the current frame, the velocity $v$ is initialized as the estimated motion $v_\text{init}$, the lifespan variance $\sigma_t$ is set to cover three frames, and the temporal radius $r$ is initialized to $1e-6$ to keep the initial temporal visibility close to the Gaussian distribution. 

For the static region, we use COLMAP~\cite{schonberger2016structure} with full images to reconstruct the first frame, which includes both dynamic and static points (the wrong dynamic points will be removed during the optimization), to initialize long-lived Gaussian primitives.
These Gaussians are initialized with zero velocity, a time $T$ corresponding to the middle of the sequence, an extended lifespan variance $\sigma_t$ lasting three times the total number of frames, and a lifespan radius initialize to $1e-6$.

\subsection{Training}
\label{training}
The dynamic scene is trained under a reconstruction loss $\mathcal{L}_{\text{recon}}$, a regularization $\mathcal{L}_{\text{reg}}$, and an auxiliary loss $\mathcal{L}_{\text{e}}$ for densification ($\mathcal{L}_{\text{e}}$ is related to $E$, which will be introduced in the supplementary material):
\begin{equation}
 \mathcal{L} = \mathcal{L}_{\text{recon}} + \mathcal{L}_{\text{reg}} + \mathcal{L}_e.
\end{equation}

\noindent\textbf{Reconstruction loss.} 
The L1 loss $\mathcal{L}_1$, SSIM loss~\cite{wang2004image} $\mathcal{L}_s$ and perceptual loss~\cite{zhang2018the} $\mathcal{L}_p$ are applied to the rendered images $\tilde{I}$ to measure the difference between the ground-truth images $I_{gt}$:
\begin{equation}
 \mathcal{L}_{\text{recon}} = \lambda_{1} \mathcal{L}_1(\tilde{I}, I_{gt}) + \lambda_{s} \mathcal{L}_s(\tilde{I}, I_{gt}) + \lambda_{p} \mathcal{L}_p(\tilde{I}, I_{gt}),
\end{equation}
in which we set $\lambda_{1}=0.8$, $\lambda_{s}=0.2$ and $\lambda_{p}=0.01$, respectively. 

\noindent\textbf{Regularization.} 
In order to further improve the quality of reconstruction, similar to PGSR~\cite{chen2024pgsr}, we introduce $\mathcal{L}_{\text{scale}}$ to make Gaussian primitives flatten as much as possible, while enhancing single view normal and depth consistency constraints with $\mathcal{L}_{\text{n}}$. All items of regularization are as follows:
\begin{equation}
 \mathcal{L}_{\text{reg}} = \lambda_{\text{scale}} \mathcal{L}_{\text{scale}} + \lambda_{\text{opacity}} \mathcal{L}_{\text{opacity}} + \lambda_{\text{n}} \mathcal{L}_{\text{n}} + \lambda_{\text{t}} \mathcal{L}_{\text{t}}.
\end{equation}
\begin{equation}
\begin{aligned}
 \mathcal{L}_{{t}} &= \frac{1}{N} \sum\frac{1}{\sqrt{-2\log(o_\text{th})\sigma_t^2}+r},\\
 \mathcal{L}_{\text{opacity}} &= \frac{1}{N} \sum O \cdot \slashed{\nabla}\!\left[l(t)\right],
 \end{aligned}
\end{equation}
where $\mathcal{L}_{\text{t}}$ extends the lifespan of Gaussian primitives, encouraging reuse of the same primitive rather than fragmenting it into multiples. $N$ is the number of Gaussian primitives, and $\slashed{\nabla}\!\left[\cdot\right]$ is the stop-gradient operation. 
%
$o_\text{th}$ denotes the truncation threshold, and Gaussian primitives with opacity $O_t<o_\text{th}$ at time $t$ are excluded from rendering.
We add the opacity loss $\mathcal{L}_{\text{opacity}}$ at the second densification stage and stop resetting opacities~\cite{kerbl3Dgaussians} to stabilize convergence.

%% file: sec_4_experiments.tex
\section{Experiments}
\begin{figure*}[ht]
    \centering
    \includegraphics[width=1.0\linewidth]{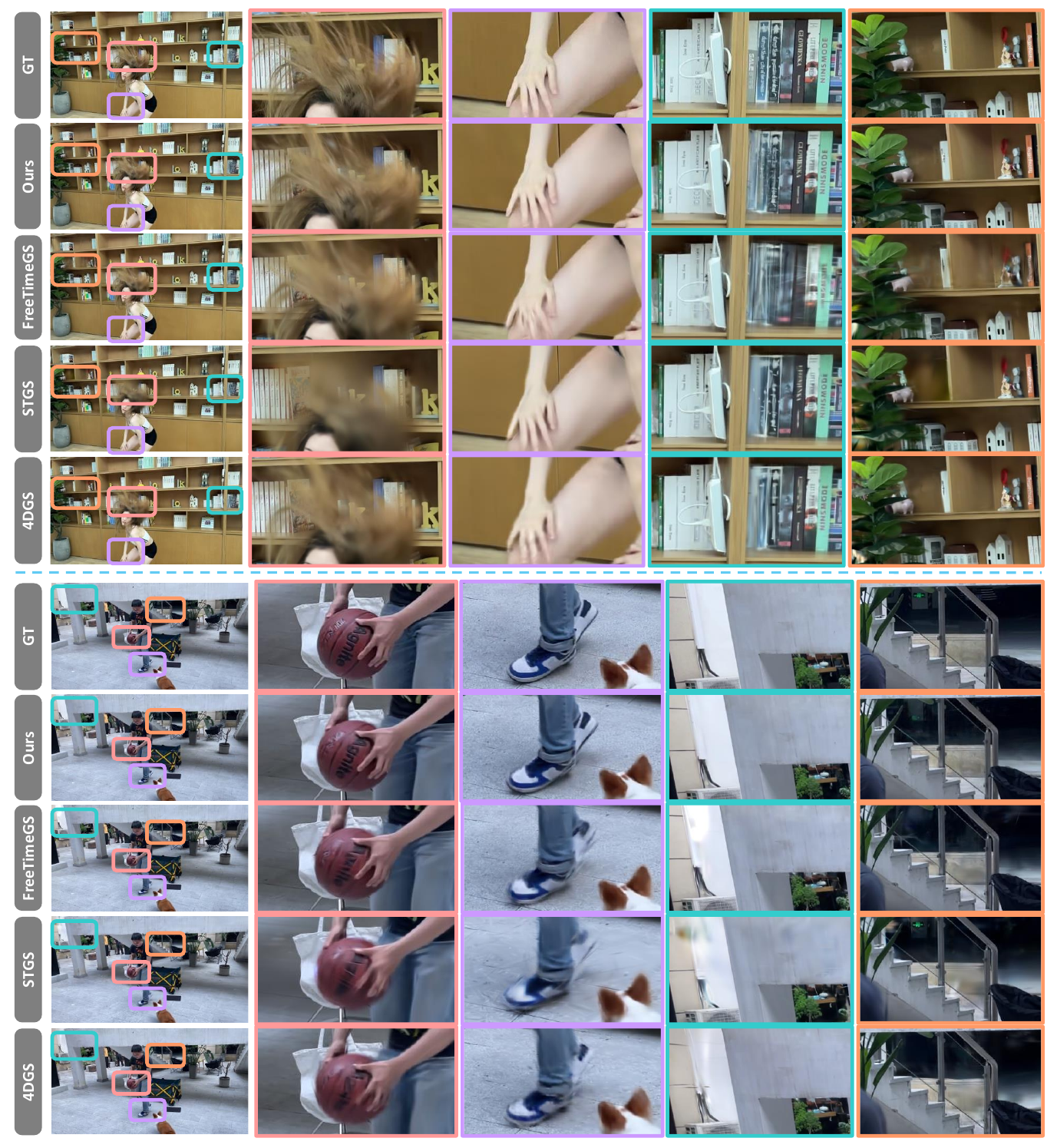}
    \vspace{-6mm}
    \caption{Qualitative comparison on the SelfCap Dataset~\cite{xu2024representing}. Our method achieves the rendering quality compared with baseline methods, especially for distant static regions (e.g., books and wall) and fast-moving dynamic regions (e.g., hairs and ball).}
    \vspace{-6mm}
    \label{fig:comparison-selfcap}
\end{figure*}
\begin{figure*}[ht]
    \centering
    \includegraphics[width=1.0\linewidth]{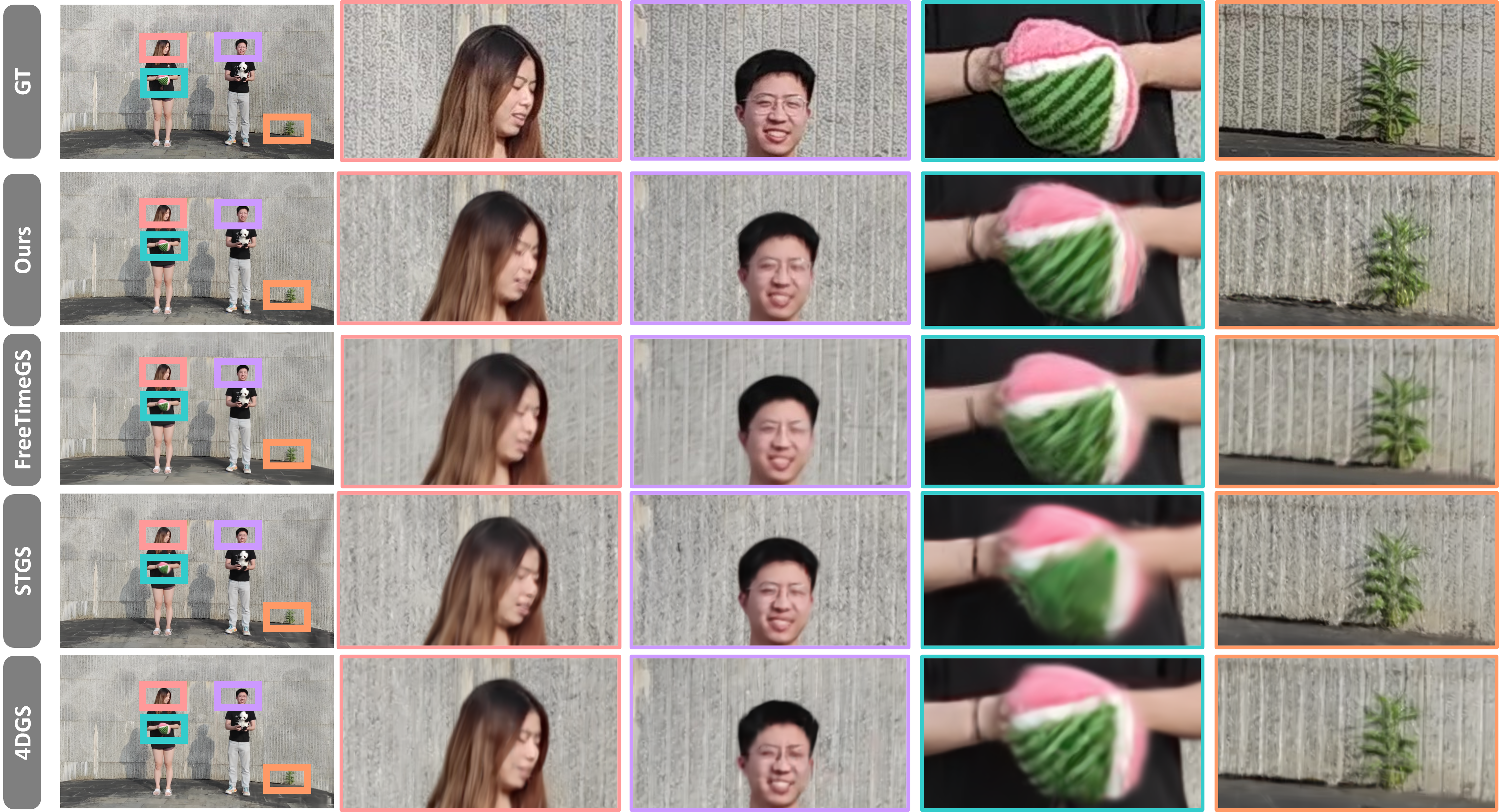}
    \caption{Qualitative comparison on the ENeRF-Outdoor Dataset~\cite{lin2022efficient}. Our method achieves the best rendering quality compared with baseline methods, especially for distant static regions and fast-moving dynamic regions.}
    \label{fig:comparison2}
\end{figure*}


\begin{figure*}[ht]
    \centering
    \includegraphics[width=1.0\linewidth]{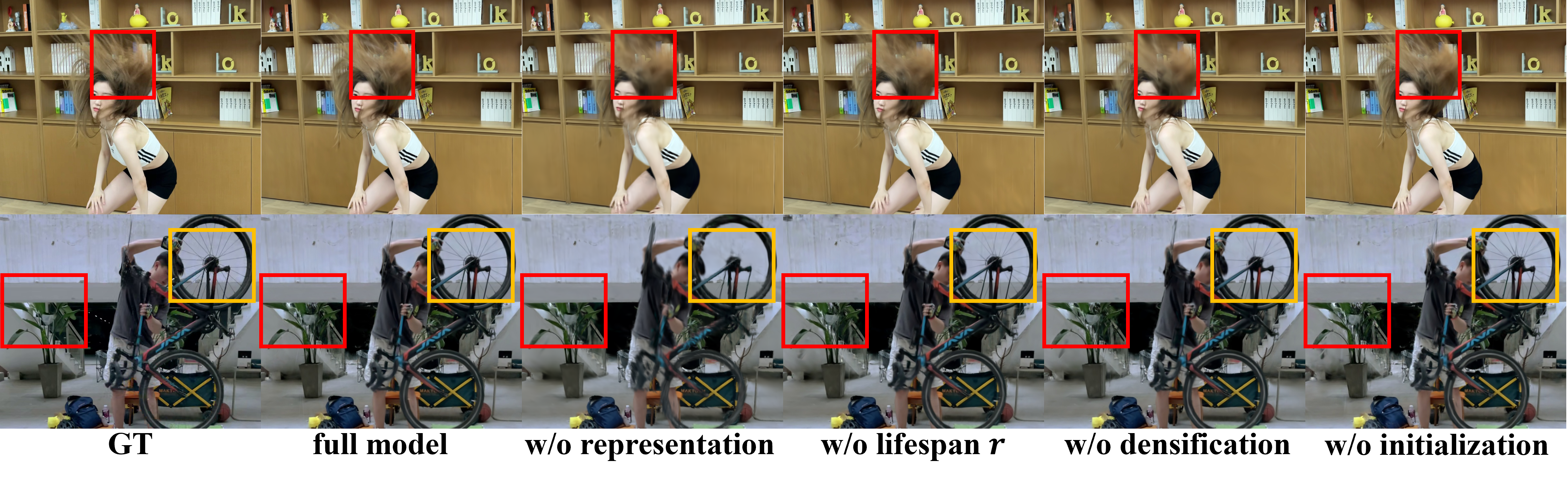}
    \vspace{-6.5mm}
    \caption{Ablation study on the SelfCap Dataset~\cite{xu2024representing}. Our full model achieves the best rendering quality, especially for distant static regions and fast-moving dynamic regions.}
    \label{fig:abl1}
    \vspace{-5mm}
\end{figure*}
\begin{table*}[h]
\centering
\vspace{-3mm}
\caption{Quantitative comparison on Neural3DV~\cite{li2022neural} Dataset, ENeRF-Outdoor~\cite{lin2022efficient} Dataset, and SelfCap~\cite{xu2024representing} Dataset. We report PSNR, $\text{SSIM}_2$~\cite{wang2004image}, and LPIPS~\cite{zhang2018the} to evaluate the rendering quality. Values in boldface denote the best result in the corresponding column. }
\label{tab:comparison}
\scalebox{0.95}{
\begin{tabular}{cccccccccc}
\hline
\multirow{2}{*}{Method} & \multicolumn{3}{c}{Neural3DV} & \multicolumn{3}{c}{ENeRF-Outdoor} & \multicolumn{3}{c}{SelfCap} \\
                        & PSNR $\uparrow$ & $\text{SSIM}_2$ $\uparrow$ & LPIPS $\downarrow$ & PSNR $\uparrow$ & $\text{SSIM}_2$ $\uparrow$ & LPIPS $\downarrow$ & PSNR $\uparrow$ & $\text{SSIM}_2$ $\uparrow$ & LPIPS $\downarrow$ \\
\hline
Deformable-3DGS~\cite{yang2024deformable} & 31.15 & 0.970 & 0.049 & 24.26 & 0.801 & 0.318 & 25.85 & 0.920 & 0.312 \\
Ex4DGS \cite{lee2024fully}                & 32.11 & 0.970 & 0.048 & 24.89 & 0.817 & 0.305 & 24.96 & 0.920 & 0.299 \\
4DGS~\cite{yang2024real}                  & 32.01 & 0.972 & 0.055 & 24.82 & 0.822 & 0.317 & 25.86 & 0.923 & 0.245\\
STGS~\cite{li2024spacetime}               & 32.05 & 0.972 & 0.044 & 24.93 & 0.818 & 0.297 & 24.77 & 0.894 & 0.291\\
FreeTimeGS~\cite{wang2025freetimegs}      & 33.19 & 0.974 & 0.036 & 25.36 & 0.846 & 0.244 & 27.50 & 0.951 & 0.201\\
\hline
Ours & \textbf{33.57} & \textbf{0.977} & \textbf{0.031} & \textbf{25.82} & \textbf{0.872} & \textbf{0.233} & \textbf{28.14} & \textbf{0.960} & \textbf{0.192}\\
\hline
\end{tabular}}
\vspace{-5mm}
\end{table*}
\subsection{Experimental Settings}
\paragraph{Datasets.} 
We evaluate our method on three widely used dynamic-scene benchmarks: Neural3DV~\cite{li2022neural}, ENeRF-Outdoor~\cite{lin2022efficient}, and SelfCap~\cite{wang2025freetimegs}.
Neural3DV contains six indoor scenes captured by 19–21 cameras at a resolution of $2704 \times 2028$ and 30 FPS. Following~\cite{wang2025freetimegs}, we use the first 300 frames of each scene and downsample images by a factor of 0.5.
ENeRF-Outdoor~\cite{lin2022efficient} provides three outdoor sequences recorded with 18 synchronized cameras at $1920 \times 1080$ and 60 FPS. We use the first 300 frames without resizing.
SelfCap~\cite{wang2025freetimegs} contains fast-motion sequences; we use six scenes (60 frames each), resizing images by 0.5 except for the bike scene, which is kept at full resolution.

%
\paragraph{Metrics.} 
We evaluate rendering quality using three widely adopted image-based metrics: Peak Signal-to-Noise Ratio (PSNR), Structural Similarity Index (SSIM)~\cite{wang2004image}, and Learned Perceptual Image Patch Similarity (LPIPS)~\cite{zhang2018the}.
PSNR and SSIM measure pixel-level fidelity and structural consistency, while LPIPS assesses perceptual similarity using deep features, providing a more human-aligned evaluation of visual quality.

\paragraph{Baselines.}
We compare our method with well-established dynamic 4D Gaussian baselines, including the canonical-deformation based method Deformable-3DGS~\cite{yang2024deformable} and motion-based approaches such as Ex4DGS~\cite{lee2024fully}, 4DGS~\cite{yang2024real}, STGS~\cite{li2024spacetime}, and FreeTimeGS~\cite{wang2025freetimegs}. For FreeTimeGS, we reproduce its pipeline for qualitative comparison, while quantitative results are taken directly from the original paper.
\subsection{Results and Comparisons}
Quantitative results on Neural3DV, ENeRF-Outdoor and SelfCap datasets are reported in Tab.~\ref{tab:comparison}. Our method consistently achieves the best performance across all metrics on all three benchmarks. 
Qualitative comparisons (Fig.~\ref{fig:comparison2} and Fig.~\ref{fig:comparison-selfcap}, and the supplementary material) further demonstrate high-fidelity reconstructions with sharper dynamics and better-preserved static details. 

Across datasets, baseline methods face a recurring trade-off in optimization, making it difficult to balance static background fidelity and dynamic reconstruction quality. 
Due to its fully coupled representation, 4DGS~\cite{yang2024real} often struggles to converge for rapidly moving content (e.g., \emph{ball} in SelfCap and \emph{watermelon} in ENeRF-Outdoor) leading to artifacts and blurred details. 
Similarly, the complex high-order formulation of STGS~\cite{li2024spacetime} complicates optimization and prevents full convergence (e.g., \emph{face} and \emph{toy} in ENeRF-Outdoor).
FreeTimeGS~\cite{wang2025freetimegs} overlooks the dependency between velocity and lifespan, yielding an overly unconstrained parameterization that degrades static fidelity (e.g., \emph{wall} in ENeRF-Outdoor and \emph{books} in SelfCap) and introduces an optimization imbalance between static and dynamic regions, leaving dynamic elements under-converged (e.g., \emph{toy} in ENeRF-Outdoor and \emph{hair/skin/ball} in SelfCap). 
In contrast, our method employs lifespan modulation to retain the strong static modeling capability of 3DGS~\cite{kerbl3Dgaussians}, thereby preserving high-fidelity backgrounds. Meanwhile, velocity-aware initialization together with velocity-lifespan aware densification improves convergence and allocates sufficient capacity to dynamic regions, enabling robust reconstruction of fast and complex motions across all benchmarks.
Moreover, the extended duration of the static segment maintains a stable background without noticeable flicker.
The specific temporal results and the free-viewpoint video can be found in the supplementary video.
\begin{table}[t]
    \centering
    \caption{Ablation study on SelfCap~\cite{xu2024representing} Dataset (Partial). We report PSNR, $\text{SSIM}_2$, and LPIPS to evaluate the rendering quality.}
    \vspace{-1mm}
    \begin{tabular}{@{}lcccccc@{}}
        \toprule
        Method & PSNR $\uparrow$ & $\text{SSIM}_2$ $\uparrow$ & LPIPS $\downarrow$\\ \midrule
        w/o our representation & 25.96 & 0.907 & 0.299\\
        w/o lifespan $r$ & 26.76 & 0.927 & 0.321\\
        w/o our densification& 26.82 & 0.919 & 0.317\\
        w/o our initialization& 26.83 & 0.927 & 0.297\\
        full model & \textbf{27.36} & \textbf{0.947} & \textbf{0.244}\\ 
        \bottomrule
    \end{tabular}
    \vspace{-4.6mm}
    \label{tab:abl}
\end{table}



\subsection{Ablation Studies}
To verify the effectiveness of our 4D representation and our densification strategy, we conduct independent experiments for each component while keeping the others unchanged. 
We then evaluate the quantitative metrics and present the qualitative results. As shown in Tab.~\ref{tab:abl}, the performance decreases when removing any of the modules we proposed.


\noindent\textbf{Effectiveness of our 4D representation.} 
To evaluate our 4D representation, we replace it with the coupled representation used in 4DGS~\cite{yang2024real} and report the results in Fig.~\ref{fig:abl1} and Tab.~\ref{tab:abl}.
The coupled baseline exhibits artifacts on fast-moving thin structures as well as in background regions.
Since motion and appearance are entangled in the coupled parameterization, optimization becomes unstable under rapid motion, which leads to local non-convergence and produces artifacts (e.g., hairs and bicycle spokes).
With decoupled 4D representation, artifacts are largely suppressed, yielding sharper dynamic details and cleaner static regions.

\noindent\textbf{Effectiveness of our temporal visibility representation.} 
To evaluate our temporal visibility representation, we use the original temporal visibility representation and compare the results in Fig.~\ref{fig:abl1} and Tab.~\ref{tab:abl}. 
Using the baseline visibility leads to temporal blur on moving details (e.g., hairs) and oversmoothing on long-lived structures (e.g., potted plant).
Compared to the smooth Gaussian-shaped visibility in the baseline, our flat-top visibility profile with steep falloff at the lifespan boundaries reduces temporal mixing and yields sharper dynamic details and quasi-static regions.


\noindent\textbf{Effectiveness of our densification.} 
To verify the effectiveness of velocity-lifespan-aware densification, we design an experiment in which we use the 4DGS~\cite{yang2024real} densification and compare results in Fig.~\ref{fig:abl1} and Tab.~\ref{tab:abl}. Because the original densification in 4DGS~\cite{yang2024real} does not account for lifespan and velocity, Gaussian primitives undergo an uneven number of updates across fast/slow and short-/long-lived regions, leading to imbalanced optimization. Fast, short-lived components are prone to non-convergence, whereas slow, long-lived components tend to overfit. 
Consequently, artifacts arise across all cases.
In contrast, our densification adapts the update frequency based on velocity and lifespan, ensuring that fast, short-lived regions are sufficiently optimized while slow, long-lived regions avoid overfitting.

\noindent\textbf{Effectiveness of velocity-aware initialization.} 
To verify the effectiveness of velocity-aware initialization, we design an experiment in which we do not separate the static and dynamic region and only use point cloud to initialize the Gaussian primitive parameter without velocity (i.e., the velocity is set to $0$). We compare the results in Fig.~\ref{fig:abl1} and Tab.~\ref{tab:abl}. 
%
It can be seen that both dynamic and static regions have artifacts (e.g., hairs and potted plant).
Therefore, our velocity-aware initialization can improve the ability to model the static region.

%% file: sec_5_conclusion.tex
\section{Discussion}
We introduced SharpTimeGS, a unified 4D Gaussian representation driven by a learnable lifespan that modulates motion and temporal visibility.
This design stabilizes long-lived primitives, preserves high-frequency dynamic motion, and eliminates motion dragging through a flat-top temporal kernel.
Combined with lifespan–velocity–aware densification and velocity-guided initialization, SharpTimeGS achieves sharper reconstructions and superior temporal consistency.
Experiments on diverse dynamic-scene benchmarks confirm clear improvements over prior 4D Gaussian methods, while retaining real-time rendering efficiency.
%

\noindent\textbf{Limitation.} 
%
%
Our reconstruction system is currently not real-time, requiring several hours to convert multi-view videos into the proposed 4D representation. 
Future improvements could focus on accelerating training through stronger geometric priors or regularizing the spatial distribution of Gaussian primitives. 
Moreover, the current representation targets novel view synthesis and does not yet support relighting.
However, since our method provides a strong geometric representation, future work can readily enable relighting by incorporating additional material and reflectance properties.